\documentclass{article}

\usepackage{microtype}
\usepackage{graphicx}
\usepackage{subcaption}
\usepackage{booktabs} 
\usepackage{amscd}
\usepackage{array}
\usepackage{amsthm}
\usepackage{amssymb}
\usepackage{amsmath}
\usepackage{amsfonts}
\usepackage{bm}  
\usepackage[algo2e]{algorithm2e} 
\usepackage[OT1]{fontenc} 
\usepackage{microtype}
\usepackage[titletoc,toc,title]{appendix}

\usepackage{hyperref}

\newtheorem{proposition}{Proposition}

\usepackage[accepted]{icml2019}

\newcommand{\bb}{\boldsymbol{b}}
\newcommand{\bc}{\boldsymbol{c}}

\newcommand{\x}{\boldsymbol{x}}

\newcommand{\w}{\boldsymbol{w}}

\newcommand{\W}{\boldsymbol{W}}

\newcommand{\bdelta}{\boldsymbol{\delta}}

\newcommand{\xpj}{\x_p^{(j)}}
\newcommand{\phixpj}{\phi(\x_p^{(j)})}
\newcommand{\phiixpj}{\phi^{(i)}(\x_p^{(j)})}
\newcommand{\phixt}{\phi(\x_t)}

\newcommand{\sumj}{\sum_{j=1}^k}

\icmltitlerunning{Transferable Clean-Label Poisoning Attacks on Neural Nets}

\begin{document}

\twocolumn[
\icmltitle{Transferable Clean-Label Poisoning Attacks on Deep Neural Nets}



\icmlsetsymbol{equal}{*}

\begin{icmlauthorlist}
\icmlauthor{Chen Zhu}{equal,umd}
\icmlauthor{W. Ronny Huang}{equal,umd}
\icmlauthor{Ali Shafahi}{umd}
\icmlauthor{Hengduo Li}{umd}
\icmlauthor{Gavin Taylor}{usna}
\icmlauthor{Christoph Studer}{cornell}
\icmlauthor{Tom Goldstein}{umd}
\end{icmlauthorlist}

\icmlaffiliation{umd}{University of Maryland, College Park}
\icmlaffiliation{usna}{United States Naval Academy}
\icmlaffiliation{cornell}{Cornell University}

\icmlcorrespondingauthor{Chen Zhu}{chenzhu@cs.umd.edu}
\icmlcorrespondingauthor{W. Ronny Huang}{wronnyhuang@gmail.com}
\icmlcorrespondingauthor{Tom Goldstein}{tomg@cs.umd.edu}

\icmlkeywords{machine learning, data poisoning, poisoning attacks, adversarial, ICML}

\vskip 0.3in
]



\printAffiliationsAndNotice{\icmlEqualContribution} 

\begin{abstract}
Clean-label poisoning attacks inject innocuous looking (and ``correctly'' labeled) poison images into training data, causing a model to misclassify a targeted image after being trained on this data.
We consider {\em transferable} poisoning attacks that succeed without access to the victim network's outputs, architecture, or (in some cases) training data. 
To achieve this, we propose a new ``polytope attack'' in which poison images are designed to surround the targeted image in feature space. 
We also demonstrate that using Dropout during poison creation helps to enhance transferability of this attack.
We achieve transferable attack success rates of over 50\% while poisoning only 1\% of the training set.
  
\end{abstract}

\section{Introduction}

Deep neural networks require large datasets for training and hyper-parameter tuning.  As a result, many practitioners turn to the web as a source for data, where one can automatically scrape large datasets with little human oversight.   
Unfortunately, recent results have demonstrated that these data acquisition processes can lead to security vulnerabilities. In particular, retrieving data from untrusted sources makes models vulnerable to  {\em data poisoning attacks} wherein an attacker injects maliciously crafted examples into the training set in order to hijack the model and control its behavior.

This paper is a call for attention to this security concern. We explore effective and transferable {\em clean-label poisoning attacks} on image classification problems. In general, data poisoning attacks aim to control the model's behavior during inference by modifying its training data~\citep{shafahi2018poison, suciu2018does,koh2017understanding, mahmoody2017}. In contrast to evasion attacks~\citep{biggio2013evasion, szegedy2013intriguing, goodfellow2015explaining} and recently proposed backdoor attacks~\citep{liu2017trojaning, Chen2017, turner2019cleanlabel}, 
we study the case where the targeted samples are {\em not} modified during inference.

Clean-label poisoning attacks differ from other poisoning attacks~\citep{biggio2012poisoning, steinhardt2017certified} in a critical way: they do not require the user to have any control over the labeling process. 
Therefore, the poison images need to maintain their malicious properties even when labeled correctly by an expert. 
Such attacks open the door for a unique threat model in which the attacker poisons datasets simply by placing malicious images on the web, and waiting for them to be harvested by web scraping bots, social media platform operators, or other unsuspecting victims. Poisons are then properly categorized by human labelers and used during training.
Furthermore, targeted clean label attacks do not indiscriminately degrade test accuracy but rather target misclassification of specific examples, rendering the presence of the attack undetectable by looking at overall model performance.

Clean-label poisoning attacks have been demonstrated only in the white-box setting where the attacker has complete knowledge of the victim model, and uses this knowledge in the course of crafting poison examples~\citep{shafahi2018poison, suciu2018does}. Black-box attacks of this type have not been explored; thus, we aim to craft clean-label poisons which transfer to unknown (black-box) deep image classifiers.

It has been demonstrated in evasion attacks that with only query access to the victim model, a substitute model can be trained to craft adversarial perturbations that fool the victim to classify the perturbed image into a specified class~\citep{papernot2017practical}. Compared to these attacks, transferable poisoning attacks remain challenging for two reasons. First, the victim's decision boundary trained on the poisoned dataset is more unpredictable than the unknown but fixed decision boundary of an evasion attack victim. Second, the attacker cannot depend on having direct access to the victim model (i.e. through queries) and must thus make the poisons model-agnostic. The latter also makes the attack more dangerous since the poisons can be administered in a distributed fashion (e.g. put on the web to be scraped), compromising more than just one particular victim model.

Here, we demonstrate an approach to produce transferable clean-label targeted poisoning attacks. We assume the attacker has no access to the victim's outputs or parameters, but is able to collect a similar training set as that of the victim. The attacker trains substitute models on this training set, and optimizes a novel objective that forces the poisons to form a polytope in feature space that entraps the target inside its convex hull. A classifier that overfits to this poisoned data will classify the target into the same class as that of the poisons. 
This new objective has better success rate than feature collision~\citep{shafahi2018poison} in the black-box setting, and it becomes even more powerful when enforced in multiple intermediate layers of the network, showing high success rates in both transfer learning and end-to-end training contexts. 
We also show that using Dropout when crafting the poisons improves transferability.

\section{The Threat Model}
Like the poisoning attack of~\citep{shafahi2018poison}, the attacker in our setting injects a small number of perturbed samples (whose labels are true to their class) into the training set of the victim. The attacker's goal is to cause the victim network, once trained, to classify a test image (not in the training set) as a specified class. 
We consider the case of image classification, where the attacker achieves its goal by adding adversarial perturbations $\bdelta$ to the images.
$\bdelta$ is crafted so that the perturbed image $\x+\bdelta$ shares the same class as the clean image $\x$ for a human labeler, while being able to change the decision boundary of the DNNs in a certain way.

Unlike~\citep{shafahi2018poison} or~\citep{papernot2017practical}, which requires full or query access to the victim model, here we assume the victim model is not accessible to the attacker, which is a practical assumption in many systems such as autonomous vehicles and surveillance systems.
Instead, we need the attacker to have knowledge about the victim's training distribution, such that a similar training set can be collected for training substitute models.  

We consider two learning approaches that the victim may adopt. 
The first learning approach is {\em transfer learning}, in which a pre-trained but frozen feature extractor $\phi$, e.g., the convolutional layers of a ResNet~\citep{he2016deep} trained on a reference dataset like CIFAR or ImageNet, is applied to images, 
and an application-specific linear classifier with parameters $\W,\bb$ is fine-tuned on the features $\phi(\mathcal{X})$ of another dataset $\mathcal{X}$. 
Transfer learning of this type is common in industrial applications when large sets of labeled data are unavailable for training a good feature extractor. 
Poisoning attacks on transfer learning were first studied in the white-box settings where the feature extractor is known in~\citep{koh2017understanding,shafahi2018poison}, and similarly
in~\citep{mei2015using,biggio2012poisoning}, all of which target linear classifiers over deep features.

The second learning approach is {\em end-to-end training}, where the feature extractor and the linear classifier are trained jointly. 
Obviously, such a setting has stricter requirements on the poisons than the transfer learning setting, since the injected poisons will affect the parameters of the feature extractor. 
\citep{shafahi2018poison} uses a watermarking strategy that superposes up to 30\% of the target image onto about 40 poison images, only to achieve about 60\% success rate in a 10-way classification setting.

\begin{figure}
  \includegraphics[width=.99\linewidth]{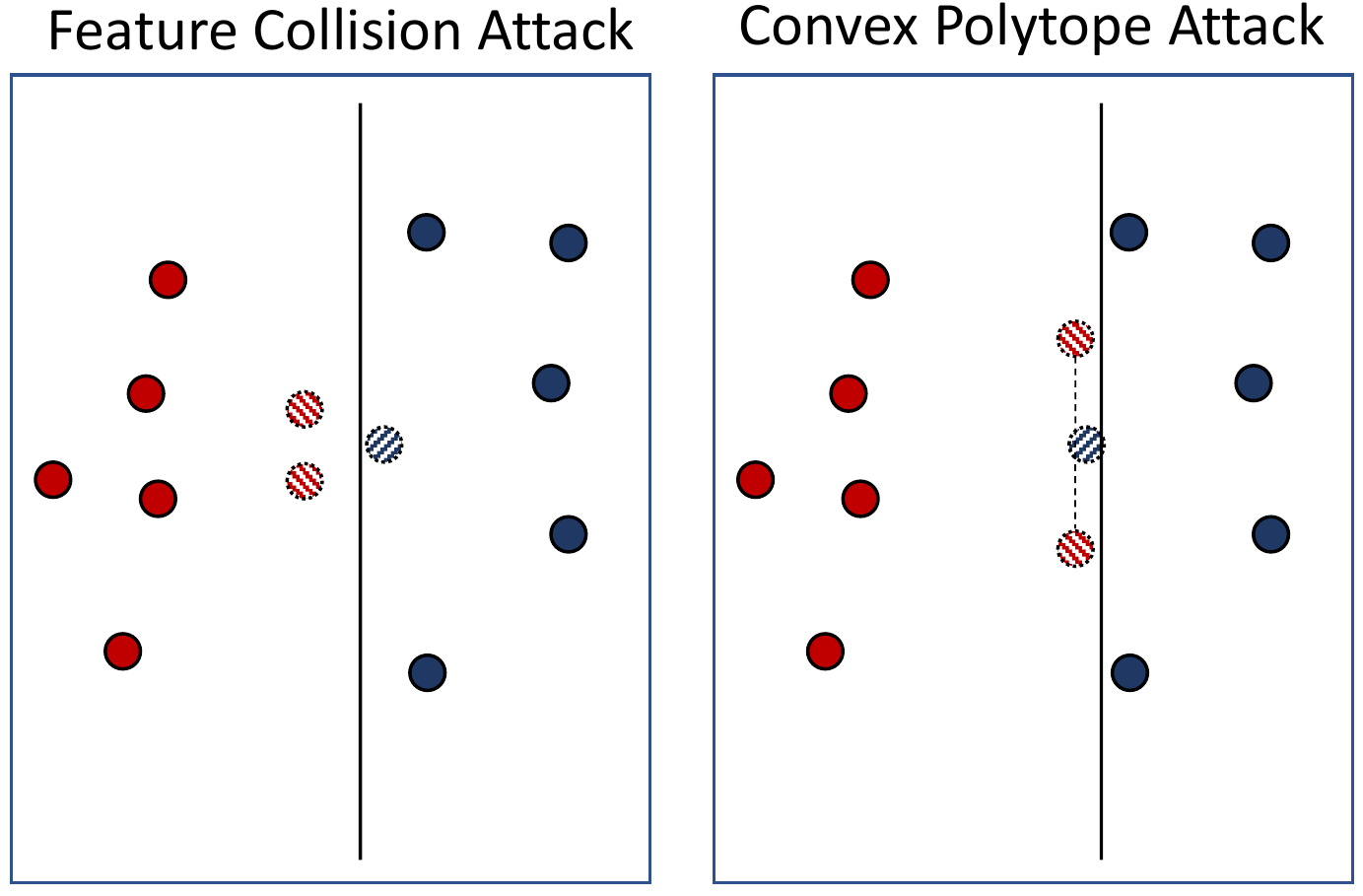}
  \caption{An illustrative toy example of a linear SVM trained on a two-dimensional space with training sets poisoned by Feature Collision Attack and Convex Polytope Attack respectively. The two striped red dots are the poisons injected to the training set, while the striped blue dot is the target, which is not in the training set. All other points are in the training set. Even when the poisons are the closest points to the target, the optimal linear SVM will classify the target correctly in the left figure. The Convex Polytope attack will enforce a small distance of the line segment formed by the two poisons to the target. When the line segment's distance to the target is minimized, the target's negative margin in the re-trained model is also minimized if it overfits.
  }
  \label{fig:fc_failure} 
\end{figure}

\section{Transferable Targeted Poisoning Attacks}
\subsection{Difficulties of Targeted Poisoning Attacks}
Targeted poisoning attacks are more difficult to pull off than targeted evasion attacks. 
For image classification, targeted evasion attacks only need to make the victim model misclassify the perturbed image into a certain class.
The model does not adjust to the perturbation, so the attacker only needs to find the shortest perturbation path to the decision boundary by solving a constrained optimization problem.
For example, in the norm-bounded white-box setting, the attacker can directly optimize $\bdelta$ to minimize the cross entropy loss $L_{CE}$ on the target label $y_{t}$ by solving $\bdelta_t = \arg\min_{||\bdelta||_{\infty} \leq \epsilon} L_{CE}(\x_t+\bdelta, y_\text{t})$ with Projected Gradient Descent~\citep{madry2017towards}.
In the norm-bounded black-box setting, with query access, the attacker can also train a substitute model to simulate the behavior of the victim via distillation, and perform the same optimization w.r.t. the substitute model to find a transferable $\bdelta$~\citep{papernot2017practical}.

Targeted poisoning attacks, however, face a more challenging problem. 
The attacker needs to get the victim to classify the target sample $\x_t$ into the alternative target class $\tilde{y}_t$ after being trained on the modified data distribution. 
One simple approach is to select the poisons from class $\tilde{y}_t$, and make the poisons as close to $\x_t$ as possible in the feature space.
A rational victim will usually overfit the training set, since it is observed in practice that generalization keeps improving even after training loss saturates~\citep{zhang2016understanding}. 
As a result, when the poisons are close to the target in the feature space, a rational victim is likely to classify $\x_t$ into $\tilde{y}_t$ since the space near the poisons are classified as $\tilde{y}_t$.
However, as shown in Figure~\ref{fig:fc_failure}, smaller distance to the target does not always lead to a successful attack. 
In fact, being close to the target might be too restrictive for successful attacks.
Indeed, there exists conditions where the poisons can be farther away from the target, but the attack is more successful.

\subsection{Feature Collision Attack}
Feature collision attacks, as originally proposed in~\citep{shafahi2018poison}, are a reliable way of producing targeted clean-label poisons on white-box models. 
The attacker selects a base example $\x_b$ from the targeted class for crafting the poisons $\x_p$, and tries to make $\x_p$ become the same as the target $\x_t$ in the feature space by adding small adversarial perturbations to $\x_b$. 
Specifically, the attacker solves the following optimization problem (\ref{makefrogs}) to craft the poisons:
 \begin{align}\label{makefrogs}
   \x_p = \arg\min_{\x} \|\x-\x_b\|^2+\mu \|\phi(\x) - \phi(\x_t)\|^2,
 \end{align}
 where  $\phi$ is a pre-trained neural feature extractor.
The first term enforces the poison to lie near the base in input space, and therefore maintains the same label as $\x_b$ to a human labeler.  
The second term forces the feature representation of the poison to collide with the feature representation of the target. 
The hyperparameter $\mu>0$ trades off the balance between these terms. 

If the poison example $\x_p$ is correctly labeled as a member of the targeted class and placed in the victim's training dataset $\chi$, then after training on $\chi$, the victim classifier learns to classify the poison's feature representation into the targeted class. 
Then, if $\x_p$'s feature distance to $\x_t$ is smaller than the margin of $\x_p$, $\x_t$ will be classified into the same class as $\x_p$ and the attack is successful. 
%
Sometimes more than one poison image is used to increase the success rate. 

Unfortunately for the attacker, different feature extractors $\phi$ will lead to different feature spaces, and a small feature space distance $d_{\phi}(\x_p, \x_t) = \|\phi(\x_p) - \phi(\x_t)\|$ for one feature extractor does not guarantee a small distance for another.
%

Fortunately, the results in~\citep{tramer2017space} have demonstrated that for each input sample, there exists an adversarial subspace for different models trained on the same dataset, such that a moderate perturbation can cause the models to misclassify the sample, which indicates it is possible to find a small perturbation to make the poisons $\x_p$ close to the target $\x_t$ in the feature space for different models, as long as the models are trained on the same dataset or similar data distributions.

With such observation, the most obvious approach to forge a black-box attack is to optimize a set of poisons $\{\x_p^{(j)}\}_{j=1}^k$ to produce feature collisions for an  {\em ensemble} of models $\{\phi^{(i)}\}_{i=1}^{m}$, where $m$ is the number of models in the ensemble.
Such a technique was also used in black-box evasion attacks~\citep{liu2017delving}.
Because different extractors produce feature vectors with different dimensions and magnitudes, we use the following normalized feature distance to prevent any one network from dominating the objective due to such biases:
\begin{equation}\label{eq:efc}
 L_\textit{FC}= \sum_{i=1}^m \sum_{j=1}^k \frac{\lVert \phi^{(i)}(\x_p^{(j)})- \phi^{(i)}(\x_t) \rVert^2} {\lVert \phi^{(i)}(\x_t) \rVert^2}.
\end{equation}



\subsection{Convex Polytope Attack}

One problem with the feature collision attack is the emergence of obvious patterns of the target in the crafted perturbations. 
Unlike prevalent objectives for evasion attacks which maximize single-entry losses like cross entropy, feature collision~\citep{shafahi2018poison} enforces each entry of the poison's feature vector $\phi(\x_p)$ to be close to $\phi(\x_t)$,
which usually results in hundreds to thousands of constraints on each poison image $\x_p$.
What is worse, in the black-box setting as Eq.~\ref{eq:efc}, the poisoning objective forces a collision over an ensemble of $m$ networks, which further increases the number of constraints on the poison images.
With such a large number of constraints, the optimizer often resorts to pushing the poison images in a direction where obvious patterns of the target will occur, therefore making $\x_p$ look like the target class.
As a result, human workers will notice the difference and take actions. 
Figure~\ref{fig:fishes} shows a qualitative example in the process of crafting poisons from images of \emph{hook} to attack a \emph{fish} image with Eq.~\ref{eq:efc}. 
Elements of the target image that are evident in the poison images include the fish's tail in the top image and almost a whole fish in the bottom image in column 3.

\begin{figure*}
  \centering
  \includegraphics[width=.95\textwidth]{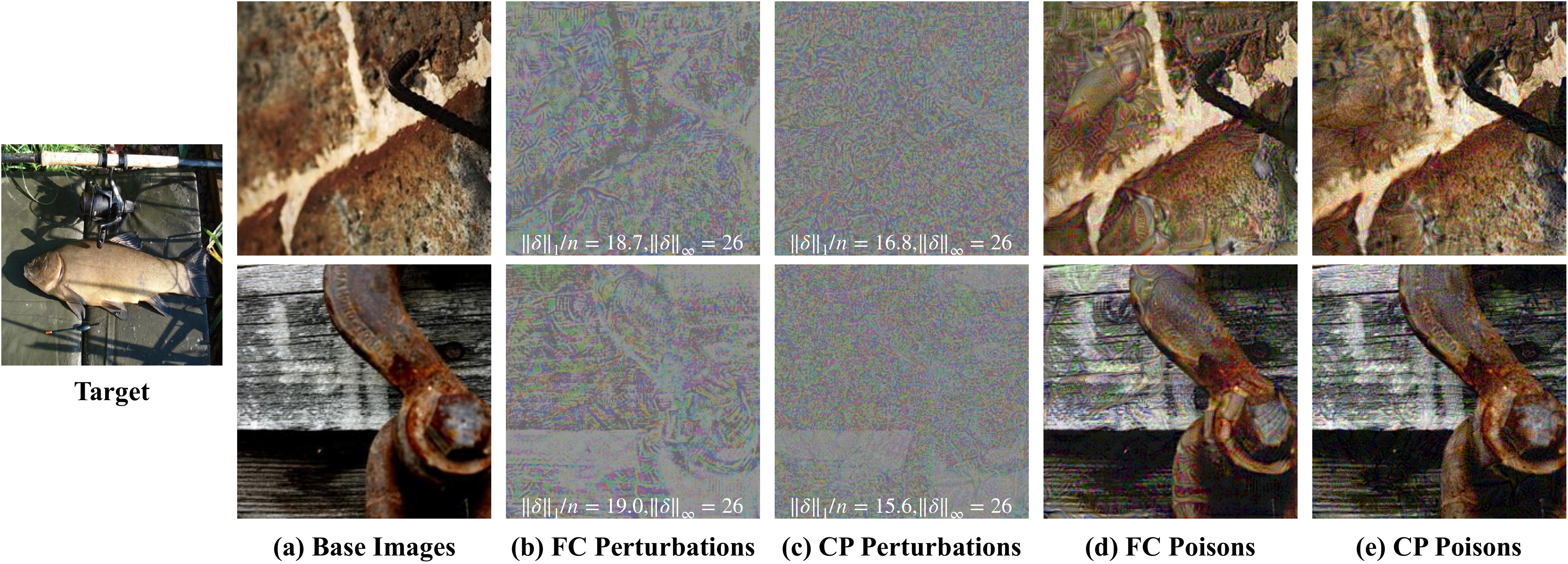}
  \caption{A qualitative example of the difference in poison images generated by Feature Collision (FC) Attack and Convex Polytope (CP) Attack. Both attacks aim to make the model mis-classify the target \emph{fish} image on the left into a \emph{hook}. 
  We show two of the five \emph{hook} images that were used for the attack, along with their perturbations and the poison images here. 
  Both attacks were successful, but unlike FC, which demonstrated strong regularity in the perturbations and obvious fish patterns in the poison images, CP tends to have no obvious pattern in its poisons.
  More details are provided in the supplementary.}
  \label{fig:fishes}
\end{figure*}

Another problem with Feature Collision Attack is its lack of transferability.
Feature Collision Attack tends to fail in the black-box setting because it is difficult to make the poisons close to $\x_t$ for every model in the feature space.
The feature space of different feature extractors should be very different, since neural networks are using one linear classifier to separate the deep features $\phi(\x)$, which has a unique solution if $\phi(\x)$ is given, but different networks usually have different accuracies.
Therefore, even if the poisons $\x_p$ collide with $\x_t$ in the feature space of the substitute models, they probably do not collide with $\x_t$ in the unknown target model, due to the generalization error.
As demonstrated by Figure~\ref{fig:fc_failure}, the attack is likely to fail even when $\x_p$ has smaller distance to $\x_t$ than its intra-class samples. 
It is also impractical to ensemble too many substitute models to reduce such error.
We provide experimental results with the ensemble Feature Collision Attack defined by Eq.~\ref{eq:efc} to show it can be ineffective.

We therefore seek a looser constraint on the poisons, so that the patterns of the target are not obvious in the poison images and the requirements on generalization are reduced. 
Noticing~\citep{shafahi2018poison} usually use multiple poisons to attack one target, we start by deriving the necessary and sufficient conditions on the set of poison features $\{\phi(\x_p^{(j)})\}_{j=1}^k$ such that the target $\x_t$ will be classified into the poison's class.

\begin{proposition}\label{prop:samelabel}
The following statements are equivalent:
\begin{enumerate}
\item Every linear classifier that classifies $\{\phixpj\}_{j=1}^k$ into label $\ell_p$ will classify $\phixt$ into label $\ell_p$.
\item $\phi(\x_t)$ is a convex combination of $\{\phi(\x_p^{(j)})\}_{j=1}^{k}$, i.e., $\phi(\x_t)=\sum_{j=1}^k c_j\phi(\x_p^{(j)})$, where $c_1,\ldots,c_k\ge 0, \sum_{j=1}^k c_j=1$. 
\end{enumerate}
\end{proposition}
The proof of Proposition~\ref{prop:samelabel} is given in the supplementary material.
In words, a set of poisons from the same class is guaranteed to alter the class label of a target into theirs if that target's feature vector lies in the {\em convex polytope} of the poison feature vectors. 
We emphasize that this is a far more relaxed condition than enforcing a feature collision---it enables the poisons to lie much farther away from the target while altering the labels on a much larger region of space. 
As long as $\x_t$ lives inside this region in the unknown target model, and $\{\x_p^{(j)}\}$ are classified as expected, $\x_t$ will be classified as the same label as $\{\x_p^{(j)}\}$. 


With this observation, we optimize the set of poisons towards forming a convex polytope in the feature space such that the target's feature vector will lie within or at least close to the convex polytope.
Specifically, we solve the following optimization problem:
\begin{equation}\label{eq:orig_obj}
\hspace{-3mm}
\begin{split}
\underset{\{\bc^{(i)}\}, \{\xpj\}}{\text{minimize}}& \frac{1}{2}\sum_{i=1}^m \frac{\lVert \phi^{(i)}(\x_t)-\sumj c_j^{(i)} \phiixpj \rVert^2 }{\lVert \phi^{(i)}(\x_t) \rVert^2} \\
\text{subject to} &\sumj c_j^{(i)}=1, c_j^{(i)}\ge 0, \forall i,j,\\
~&\lVert \x_p^{(j)} - \x_b^{(j)} \rVert_{\infty} \le \epsilon, \forall j, \\[-0.45cm]
\end{split}
\end{equation}

where $\x_b^{(j)}$ is the clean image of the $j$-th poison, and $\epsilon$ is the maximum allowable perturbation such that the perturbations are not immediately perceptible. 


Eq.~\ref{eq:orig_obj} simultaneously finds a set of poisons $\{\x_p^{(j)}\}$, and a set of convex combination coefficients $\{c_j^{(i)}\}$ such that the target lies in or close to the convex polytope of the poisons in the feature space of the $m$ models. 
Notice the coefficients $\{c_j^{(i)}\}$ are untied, i.e., they are allowed to vary across different models, which does not require $\phi^{(i)}(\x_t)$ to be close to any specific point in the polytope, including the vertices $\{\x_p^{(j)}\}$.
Given the same amount of perturbation, such an objective is also more relaxed than Feature Collision Attack (Eq.~\ref{eq:efc}) since Eq.~\ref{eq:efc} is a special case of Eq.~\ref{eq:orig_obj} when we fix $c_j^{(i)}=1/k$.
As a result, the poisons demonstrate almost no patterns of the target, and the imperceptibility of the attack is enhanced compared with feature collision attack, as shown in Figure~\ref{fig:fishes}. 

The most important benefit brought by the convex polytope objective is the improved transferability. 
For Convex Polytope Attack, $\x_t$ does not need to align with a specific point in the feature space of the unknown target model. 
It only needs to lie within the convex polytope formed by the poisons. 
In the case where this condition is not satisfied, Convex Polytope Attack still has advantages over Feature Collision Attack.
Suppose for a given target model, a residual\footnote{For FC, it is $\min_j \lVert \phi^{(t)}(\x_p^{(j)}) - \phi^{(t)}(\x_t) \rVert$; for CP, it is $\lVert \sum_j c_j^{(t)}\phi^{(t)}(\x_p^{(j)}) - \phi^{(t)}(\x_t) \rVert$} smaller than $\rho$ will guarantee a successful attack\footnote{When the residual is small enough, $\phi^{(t)}(\x_t)$ will not cross the decision boundary if poisons are classified as expected.}.
For Feature Collision Attack, the target's feature needs to lie within a $\rho$-ball centered at $\phi^{(t)}(\x_p^{(j^*)}) $, where $j^*=\arg\min_{j} \lVert \phi^{(t)}(\x_p^{(j)}) - \phi^{(t)}(\x_t) \rVert $. 
For Convex Polytope Attack, the target's feature could lie within the $\rho$-expansion of the convex polytope formed by $\{ \phi^{(t)}(\x_p^{(j)})\}_{j=1}^k $, which has a larger volume than the aforementioned $\rho$-ball, and thus tolerates larger generalization error.

\subsection{An Efficient Algorithm for Convex Polytope Attack}
We optimize the non-convex and constrained problem \eqref{eq:orig_obj} using an alternating method that side-steps the difficulties posed by the complexity of $\{\phi^{(i)}\}$ and the convex polytope constraints on $\{\bc^{(i)}\}$.
Given $\{\xpj\}_{j=1}^k$, we use forward-backward splitting~\citep{goldstein2014field} to find the the optimal sets of coefficients $\{\bc^{(i)}\}$. 
This step takes much less computation than back-propagation through the neural network, since the dimension of $\bc^{(i)}$ is usually small (in our case a typical value is 5).
Then, given the optimal $\{\bc^{(i)}\}$ with respect to $\{\xpj\}$, we take one gradient step to optimize $\{\xpj\}$, since back-propagation through the $m$ networks is relatively expensive.
Finally, we project the poison images to be within $\epsilon$ units of the clean base image so that the perturbation is not obvious, which is implemented as a clip operation.
We repeat this process to find the optimal set of poisons and coefficients, as shown in~Algorithm~\ref{alg:opt}.

In our experiments, we find that after the first iteration, initializing $\{\bc^{(i)}\}$ to the value from the last iteration accelerates its convergence.
We also find the loss in the target network to bear high variance without momentum optimizers.
Therefore, we choose Adam~\citep{kingma2014adam} as the optimizer for the perturbations as it converges more reliably than SGD in our case.
Although the constraint on the perturbation is $\ell_\infty$ norm, in contrast to~\citep{dong2018boosting} and the common practices for crafting adversarial perturbations such as FGSM~\citep{kurakin2016adversarial}, we do not take the sign of the gradient, which further reduces the variance caused by the flipping of signs when the update step is already small.

\begin{algorithm}[H]
 \KwData{Clean base images$\{\x_b^{(j)}\}_{j=1}^k$, substitute networks $\{\phi^{(i)}\}_{i=1}^m$, and maximum perturbation $\epsilon$.}
 \KwResult{A set of perturbed poison images $\{\x_p^{(j)}\}_{j=1}^k$.}
 Initialize $\bc^{(i)}\leftarrow\frac{1}{k}\bm{1}$, $\x_p^{(j)}\leftarrow\x_b^{(j)}$\;

 \While{not converged}{
 \For{i=1,\ldots,m}{
  ~~$A \leftarrow [\phi^{(i)}(\x_p^{(1)}),\ldots,\phi^{(i)}(\x_p^{(k)})]$\\ 
  $\alpha\leftarrow 1/\lVert A^\top A \rVert_2$\;\\
 \While{not converged}{

 	$\bc^{(i)}\leftarrow \bc^{(i)} - \alpha A^\top(A\bc^{(i)}-\phi^{(i)}(\x_t))$\;

 	project $\bc^{(i)}$ onto probability simplex\;
 }}

 Gradient step on $\x_p^{(j)}$ with Adam\;
 
 Clip $\x_p^{(j)}$ so that the infinity norm constraint is satisfied.
 }
 \caption{Convex Polytope Attack}
 \label{alg:opt}
\end{algorithm}

\subsection{Multi-Layer Convex Polytope Attack}
When the victim trains its feature extractor $\phi$ in addition to the classifier (last layer), enforcing Convex Polytope Attack only on the feature space of $\phi^{(i)}$ is not enough for a successful attack as we will show in experiments.
In this setting, the change in feature space caused by a model trained on the poisoned data will also make the polytope more difficult to transfer. 

Unlike linear classifiers, deep neural networks have much better generalization on image datasets.
Since the poisons all come from one class, the whole network can probably generalize well enough to discriminate the distributions of $\x_t$ and $\x_p$, such that after trained with the poisoned dataset, it will still classify $\x_t$ correctly.
As shown in Figure~\ref{fig:compare_multilayer}, when CP attack is applied to the last layer's features, the lower capacity models like SENet18 and ResNet18 are more susceptible to the poisons than other larger capacity models.
However, we know there probably exist poisons with small perturbations that are transferable to networks trained on the poison distribution, as empirical evidence from~\citep{tramer2017space} have shown there exist a common adversarial subspace for different models trained on the same dataset, and naturally trained networks usually have large enough Lipschitz to cause mis-classification~\citep{szegedy2013intriguing}, which hopefully will also be capable of shifting the polytope into such a subspace to lie close to $\x_t$.

One strategy to increase transferability to models trained end-to-end on the poisoned dataset is to jointly apply Convex Polytope Attack to multiple layers of the network.
The deep network is broken into shallow networks $\phi_1,...,\phi_n$ by depth, and the objective now becomes
\begin{equation}\label{eq:multi_layer_obj}
\hspace{-3mm}
\underset{\{\bc^{(i)}_l\}, \{\xpj\}}{\text{minimize}} \sum_{l=1}^n\sum_{i=1}^m \frac{\lVert \phi_{1:l}^{(i)}(\x_t)-\sumj c_{l,j}^{(i)} \phi_{1:l}^{(i)}(\x_p^{(j)}) \rVert^2 }{\lVert \phi_{1:l}^{(i)}(\x_t) \rVert^2},
\end{equation}
where $\phi_{1:l}^{(i)}$ is the concatenation from $\phi_1^{(i)}$ to $\phi_l^{(i)}$.
Networks similar to ResNet are broken into blocks separated by pooling layers, and we let $\phi_{l}^{(i)}$ be the $l$-th layer of such blocks.
The optimal linear classifier trained with the features up to $\phi_{1:l}$ ($l<n$) will have worse generalization than the optimal linear classifier trained with features of $\phi$, and therefore the feature of $\x_t$ should have higher chance to deviate from the features of the same class after training, which is a necessary condition for a successful attack.
Meanwhile, with such an objective, the perturbation is optimized towards fooling models with different depths, which further increases the variety of the substitute models and adds to the transferability.

\subsection{Improved Transferability via Network Randomization}
\begin{figure}
\centering
 \includegraphics[width=0.49\linewidth]{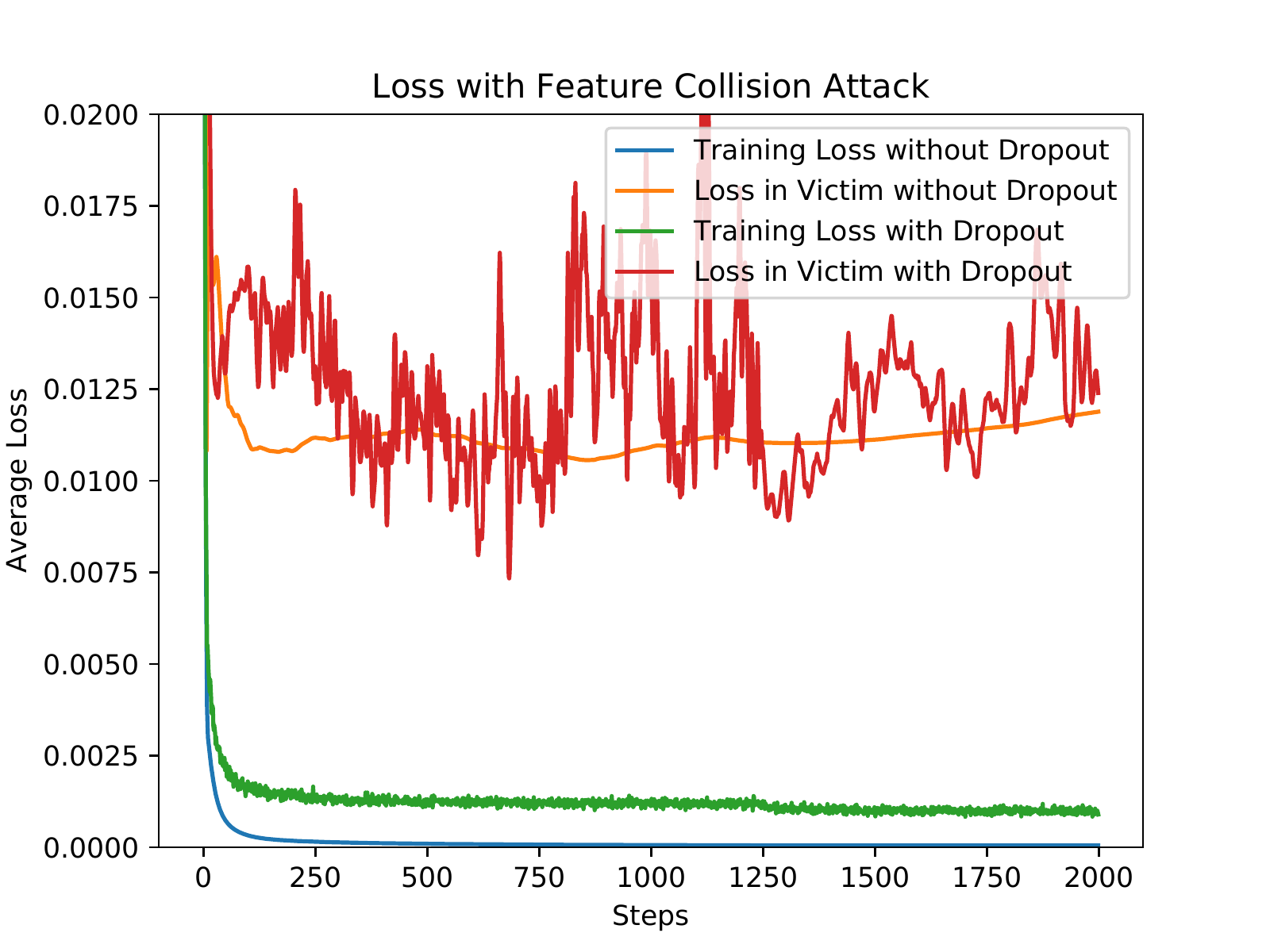}
 \includegraphics[width=0.49\linewidth]{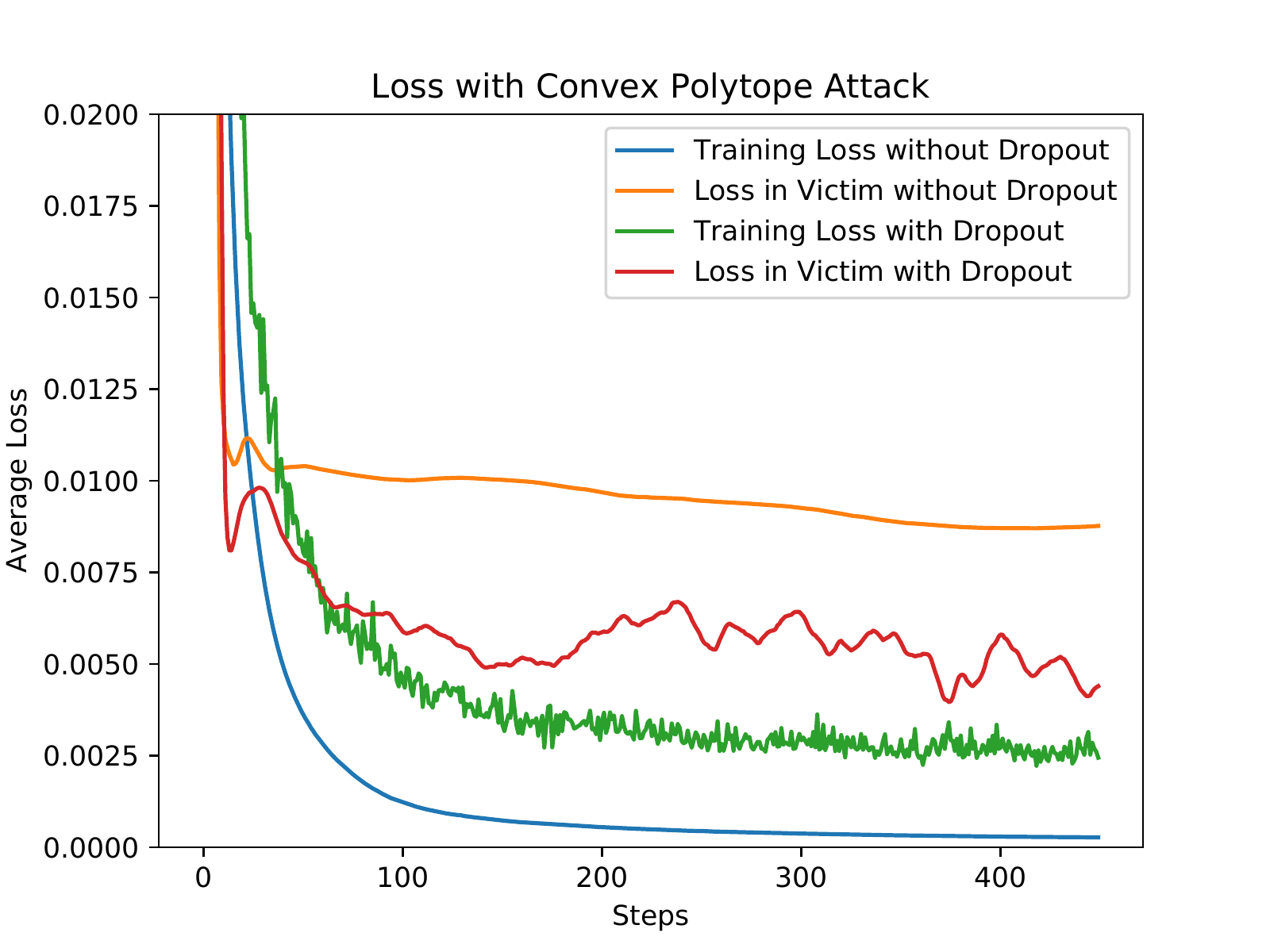}
 \caption{Loss curves of Feature Collision and Convex Polytope Attack on the substitute models and the victim models, tested using the target with index 2. Dropout improved the minimum achievable test loss for the FC attack, and improved the test loss of the CP attack significantly.}
  \vspace{-1.5em}
 \label{fig:loss_curves}
\end{figure}
Even when trained on the same dataset, different models have different accuracy and therefore different distributions of the samples in the feature space.
Ideally, if we craft the poisons with arbitrarily many networks from the function class of the target network then we should be able effectively minimize Eq.~\ref{eq:orig_obj} in the target network.
It is, however, impractical to ensemble a large number of networks due to memory constraints.

To avoid ensembling too many models, we randomize the networks with Dropout~\citep{srivastava2014dropout}, turning it on when crafting the poisons. 
In each iteration, each substitute network $\phi^{(i)}$ is randomly sampled from its function class by shutting off each neuron with probability $p$, and multiplying all the ``on'' neurons with $1/(1-p)$ to keep the expectation unchanged.
In this way, we can get an exponential (in depth) number of different networks for free in terms of memory. 
Such randomized networks increase transferability in our experiments. 
One qualititative example is given in Figure~\ref{fig:loss_curves}.
\section{Experiments}
\begin{figure}
\centering
  \includegraphics[width=0.85\linewidth]{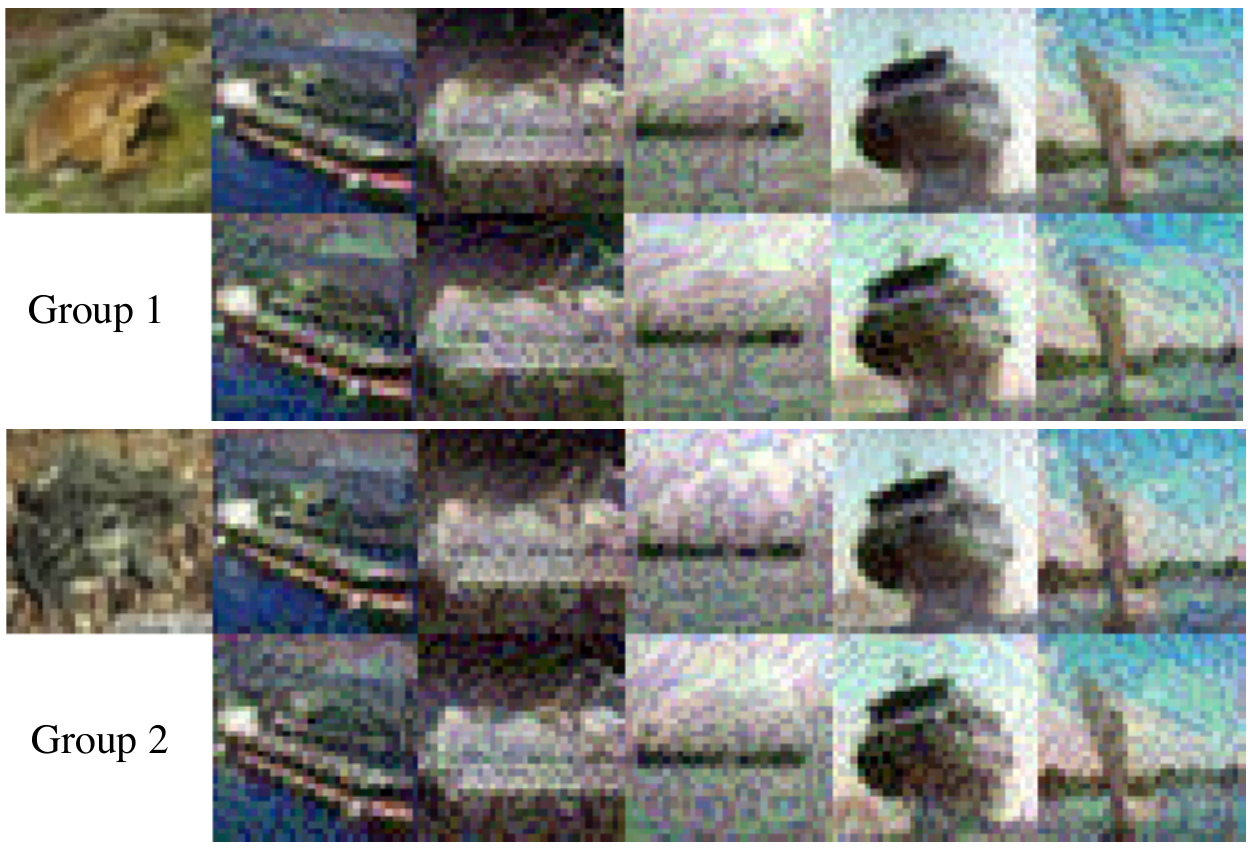}
  \caption{Qualitative results of the poisons crafted by FC and CP. 
  Each group shows the target along with five poisons crafted to attack it, where the first row is the poisons crafted with FC, and the second row is the poisons crafted with CP. 
  In the first group, the CP poisons fooled a DenseNet121 but FC poisons failed, in the second group both succeeded.
  The second target's image is more noisy and is probably an outlier of the frog class, so it is easier to attack. 
  The poisons crafted by CP contain fewer patterns of $\x_t$ than with FC, and are harder to detect. }
  \label{fig:qualitative} 
\end{figure}
In the following, we will use CP and FC as abbreviations for Convex Polytope Attacks and Feature Collision attacks respectively.
The code for the experiments is available at \url{https://github.com/zhuchen03/ConvexPolytopePosioning}.

\textbf{Datasets}~~
In this section, all images come from the CIFAR10 dataset. 
If not explicitly specified, we take the first 4800 images from each of the 10 classes (a total of 48000 images) in the training set to pre-train the victim models and the substitute models ($\phi^{(i)}$).
We leave the test set intact so that the accuracy of these models under different settings can be evaluated on the standard test set and compared directly with the benchmarks.
A successful attack should not only have unnoticeable image perturbations, but also unchanged test accuracy after fine-tuning on the clean and poisoned datasets. 
As shown in the supplementary, our attack preserves the accuracy of the victim model compared with the accuracy of those tuned on the corresponding clean dataset.

The remaining 2000 images of the training set serve as a pool for selecting the target, crafting the poisons, and fine-tuning the victim networks.
We take the first 50 images from each class (a total of 500 images) in this pool as the clean fine-tuning dataset.
This resembles the scenario where pretrained models on large datasets like Imagenet~\citep{krizhevsky2012imagenet} are fine-tuned on a similar but usually disjoint dataset. 
We randomly selected ``ship'' as the target class, and ``frog'' as the targeted image's class, i.e., the attacker wants to cause a particular frog image to be misclassified as a ship. 
The poison images $\x_p^{(j)}$ across \emph{all} experiments are crafted from the first 5 images of the ship class in the 500-image fine-tuning dataset.
We evaluate the poison's efficacy on the next 50 images of the frog class. Each of these images is evaluated independently as the target $\x_t$ to collect statistics. Again, the target images are not included in the training and fine-tuning set.

\textbf{Networks}~~
Two sets of substitute model architectures are used in this paper. 
Set \texttt{S1} includes SENet18~\citep{hu2018squeeze}, ResNet50~\citep{he2016deep}, ResNeXt29-2x64d~\citep{xie2017aggregated}, DPN92~\citep{chen2017dual}, MobileNetV2~\citep{sandler2018mobilenetv2} and GoogLeNet~\citep{szegedy2015going}. 
Set \texttt{S2} includes all the architectures of \texttt{S1} except for MobileNetV2 and GoogLeNet. \texttt{S1} and \texttt{S2} are used in different experiments as specified below.
ResNet18 and DenseNet121~\citep{huang2017densely} were used as the black-box model architectures.
The poisons are crafted on models from the set of substitute model architectures.
We evaluate the poisoning attack against victim models from the 6 different substitute model architectures as well as from the 2 black-box model architectures. Each victim model, however, was trained with different random seeds than the substitute models. If the victim's architecture appears in the substitute models, we call it a gray-box setting; otherwise, it is a black-box setting.

We add a Dropout layer at the output of each Inception block for GoogLeNet, and in the middle of the convolution layers of each Residual-like blocks for the other networks.
We train these models from scratch on the aforementioned 48000-image training set with Dropout probabilities of 0, 0.2, 0.25 and 0.3, using the same architecture and hyperparameters (except for Dropout) of a public repository\footnote{\url{https://github.com/kuangliu/pytorch-cifar}}.
The victim models that we evaluate were not trained with Dropout. 

\textbf{Attacks}~~
We use the same 5 poison ship images to attack each frog image.
For the substitute models, we use 3 models trained with Dropout probabilities of 0.2, 0.25, 0.3 from each architecture, which results in 18 and 12 substitute models for \texttt{S1} and \texttt{S2} respectively. 
When crafting the poisons, we use the same Dropout probability as the models were trained with. 
For all our experiments, we set $\epsilon=0.1$. 
We use Adam~\citep{kingma2014adam} with a relatively large learning rate of 0.04 for crafting the poisons, since the networks have been trained to have small gradients on images similar to the training set. 
We perform no more than 4000 iterations on the poison perturbations in each experiment. 
Unless specified, we only enforce Eq.~\ref{eq:orig_obj} on the features of the last layer.

For the victim, we choose its hyperparameters during fine-tuning such that it overfits the 500-image training set, which satisfies the aforementioned rational victim assumption. 
In the transfer learning setting, where only the final linear classifier is fine-tuned, we use Adam with a large learning rate of 0.1 to overfit. 
In the end-to-end setting, we use Adam with a small learning rate of $10^{-4}$ to overfit. 

\subsection{Comparison with Feature Collision}
\begin{figure}
\centering
 \includegraphics[width=0.85\linewidth]{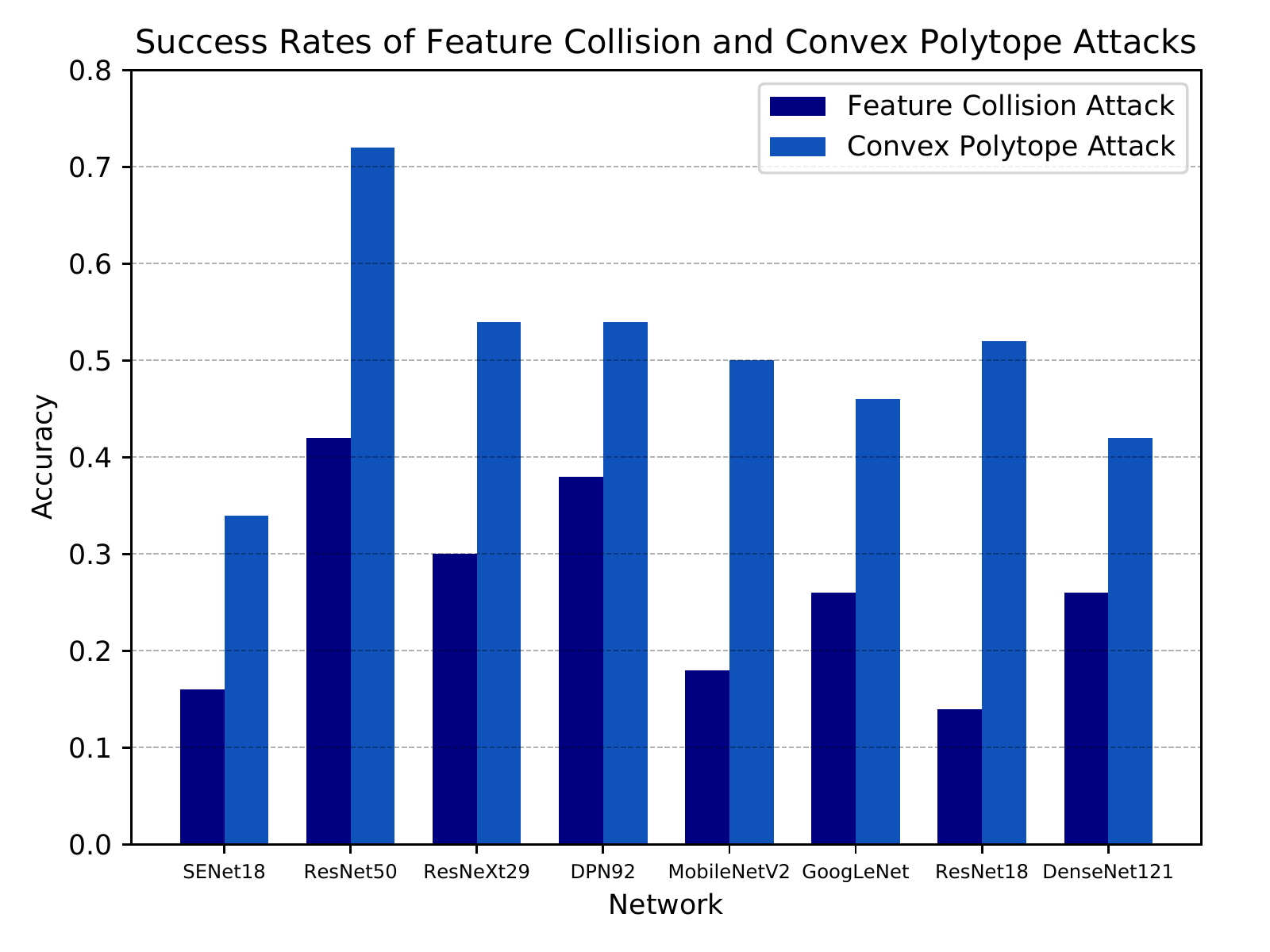}
 \caption{Success rates of FC and CP attacks on various models. Notice the first six
  entries are the gray-box setting where the models with same architecture but different weights are in
  the substitute networks, while the last two
  entries are the black-box setting.}
 \label{fig:compare_with_fc}
 \vspace{-1em}
\end{figure}
We first compare the transferability of poisons generated by FC and CP in the transfer learning training context. 
The results are shown in Figure~\ref{fig:compare_with_fc}. 
We use set \texttt{S1} of substitute architectures. FC never achieves a success rate higher than 0.5, while CP achieves success rates higher or close to 0.5 in most cases. 
A qualitative example of the poisons crafted by the two approaches is shown in Figure~\ref{fig:qualitative}.

\subsection{Importance of Training Set}
\begin{figure}
\centering
 \includegraphics[width=0.85\linewidth]{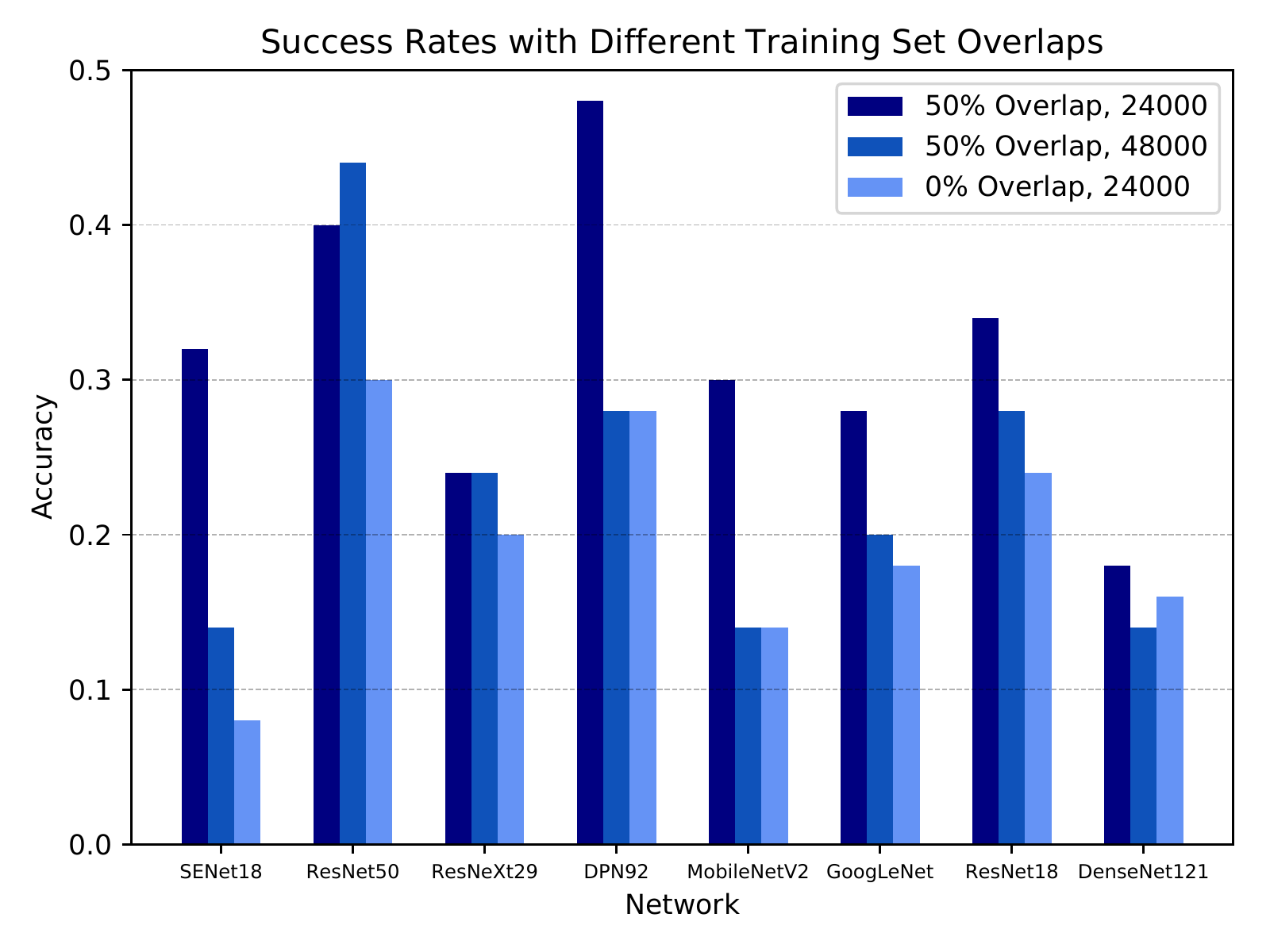}
 \caption{Success rates of Convex Polytope Attack, with poisons crafted by substitute models trained on the first 2400 images of each class of CIFAR10. The models corresponding to the three settings are trained with samples indexed from 1201 to 3600, 1 to 4800 and 2401 to 4800, corresponding to the settings of 50\%, 50\% and 0\% training set overlaps respectively. }
 \label{fig:compare_overlap}
 \vspace{-1.5em}
\end{figure}
Despite being much more successful than FC, questions remain about how reliable CP will be when we have no knowledge of the victim's training set.
In the last section, we trained the substitute models on the same training set as the victim. 
In Figure~\ref{fig:compare_overlap} we provide results for when the substitute models' training sets are similar to (but mostly different from) that of the victim. 
Such a setting is sometimes more realistic than the setting where no knowledge of the victim's training set is required, but query access to the victim model is needed~\citep{papernot2017practical},
since query access is not available for scenarios like surveillance. 
We use the less ideal \texttt{S2}, which has 12 substitute models from 4 different architectures. 
Results are evaluated in the transfer learning setting. 
Even with no data overlap, CP can still transfer to models with very different structure than the substitute models in the black-box setting.
In the 0\% overlap setting, the poisons transfer better to models with higher capacity like DPN92 and DenseNet121 than to low-capacity ones like SENet18 and MobileNetV2, probably because high capacity models overfit more to their training set. 
Overall, we see that CP may remain powerful without access to the training data in the transfer learning setting, as long as the victim's model has good generalization. 

\subsection{End-to-End Training}
\begin{figure}
\centering
 \includegraphics[width=0.85\linewidth]{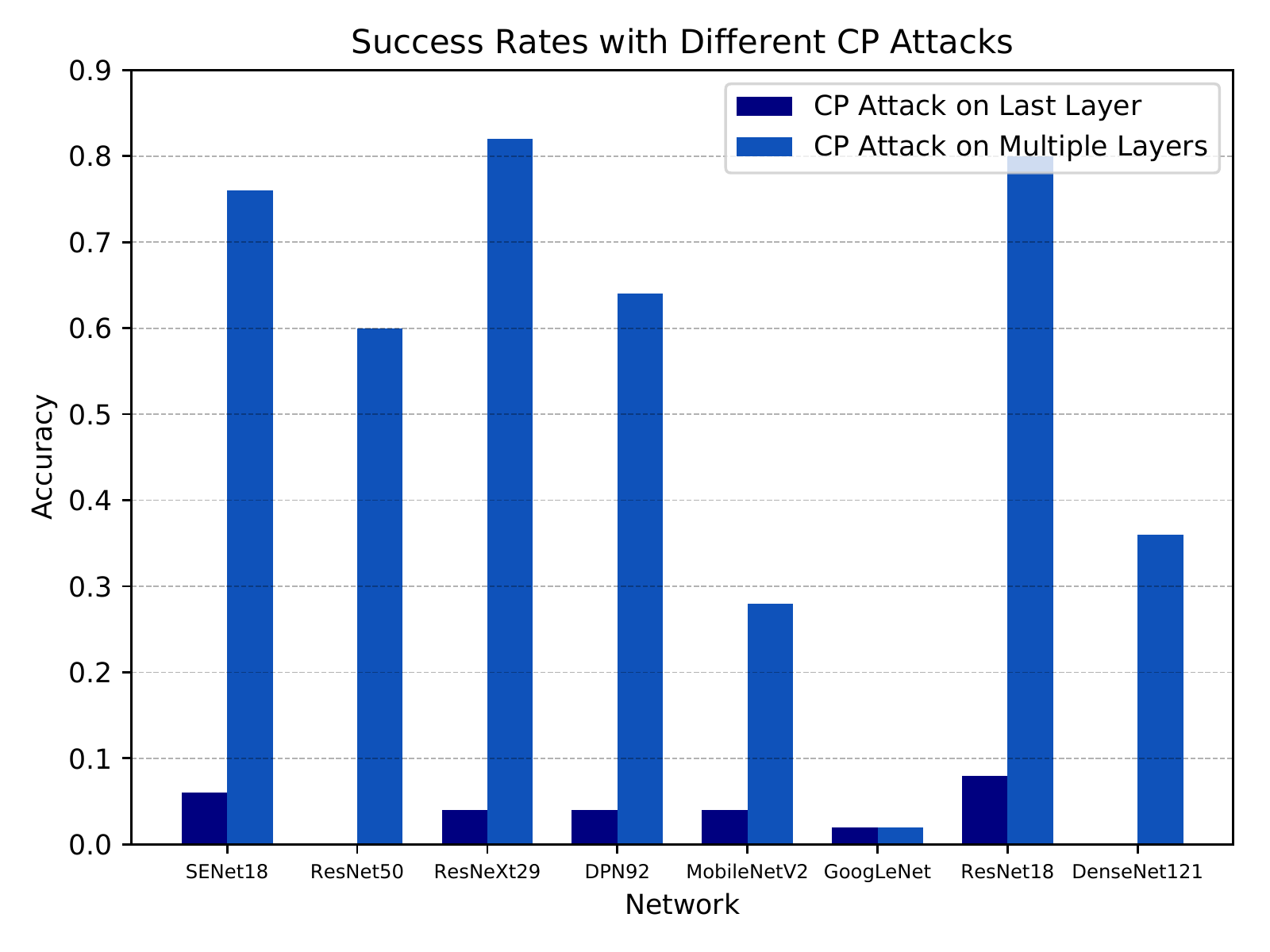}
 \caption{Success rates of Convex Polytope Attack in the end-to-end training setting. We use \texttt{S2} for the substitute models.}
 \label{fig:compare_multilayer}
 \vspace{-1.5em}
\end{figure}
A more challenging setting is when the victim adopts end-to-end training. 
Unlike the transfer learning setting where models with better generalization turn out to be more vulnerable, here good generalization may help such models classify the target correctly despite the poisons.
As shown in Figure~\ref{fig:compare_multilayer}, CP attacks on the last layer's feature is not enough for transferability, leading to almost zero successful attacks.
It is interesting to see that the success rate on ResNet50 is 0, which is an architecture in the substitute models, while the success rate on ResNet18 is the highest, which is not an architecture in the substitute models but should have worse generalization.

Therefore, we enforce CP in multiple layers of the substitute models, which breaks the models into lower capacitie ones and leads to much better results. 
In the gray-box setting, all of the attacks achieved more than 0.6 success rates. 
However, it remains very difficult to transfer to GoogLeNet, which has a more different architecture than the substitute models. 
It is therefore more difficult to find a direction to make the convex polytope survive end-to-end training.

\section{Conclusion}
In summary, we have demonstrated an approach to enhance the transferability of clean-labeled targeted poisoning attacks. 
The main contribution is a new objective which constructs a convex polytope around the target image in feature space, so that a linear classifier which overfits the poisoned dataset is guaranteed to classify the target into the poisons' class.
We provided two practical ways to further improve transferability. 
First, turn on Dropout while crafting poisons, so that the objective samples from a variety (i.e. ensemble) of networks with different structures. 
Second, enforce the convex polytope objective in multiple layers, which enables attack success even in end-to-end learning contexts.
Additionally, we found that transferability can depend on the data distribution used to train the substitute model.

\section{Acknowledgements}
Goldstein, Shafahi, and Chen were supported by the Office of Naval Research (N00014-17-1-2078), DARPA Lifelong Learning Machines (FA8650-18-2-7833), the DARPA YFA program (D18AP00055), and the Sloan Foundation. Studer was supported in part by Xilinx, Inc. and by the US National Science Foundation (NSF) under grants ECCS-1408006, CCF-1535897, CCF-1652065, CNS-1717559, and ECCS-1824379.


\appendix
\section{Proof of Proposition 1}

$2\implies 1$

For multi-class problems, the condition for $\phi(\x)$ to be classified as $\ell_p$ is 
$$\w_{\ell_p}^\top\phi(\x)+b_{\ell_p} > \w_{i}^\top\phi(\x)+b_{i}, \,\, \text{ for all } i\neq \ell_p.$$ 
Each of these constraints is linear, and is satisfied by a convex half-space.  The region that satisfies all of these constraints in an intersection of convex half-spaces, and so is convex.  Under condition (2), $\phi(\x_t)$ is a convex combination of points in this convex region, and so $\phi(\x_t)$ is itself in this convex region.

\noindent $1\implies 2$

Suppose that (1) holds.   Let $$\mathcal{S} = \{ \sum_i  c_i \phi(\x_p^j) |  \sum_i c_i=1, 0\le c_i \le 1\}$$
be the convex hull of the points $\{\phi(\x_p^j)\}_{j=1}^k.$
Let $\mathbf{u}_t = \arg\min_{\mathbf{u}\in \mathcal{S}}  \|\mathbf{u}-\phi(\x_t) \|$ be the closest point to $\phi(\x_t)$ in $\mathcal{S}.$   If $\|\mathbf{u}_t-\phi(\x_t)\|=0,$ then (2) holds and the proof is complete.  If $\|\mathbf{u}_t-\phi(\x_t)\|>0,$ then define the classifier function
  $$f(\mathbf{z}) = (\mathbf{u}_t-\phi(\x_t))^\top(  \mathbf{z}  - \mathbf{u}_t). $$ 
  Clearly $f(\phi(\x_t))<0.$ By condition (1), there is some $j$ with $f(\phi(\x_p^j))<0$ as well.  Consider the function
  $$g(\eta) = \frac{1}{2}\|  \mathbf{u}_t + \eta (\phi(\x_p^j)-\mathbf{u}_t ) -\phi(\x_t)  \|^2.$$
  Because  $\mathbf{u}_t$ is the closest point to  $\phi(\x_t)$ in $\mathcal{S},$ and $g$ is smooth,  the derivative of $g$ with respect to $\eta,$ evaluated at $\eta=0,$ is 0.  We can write this derivative condition as 
  $$g'(0) = (\mathbf{u}_t  -\phi(\x_t)  )^\top(\phi(\x_p^j)-\mathbf{u}_t )=f(\phi(\x_p^j))\ge 0.$$
  However this statement is a contradiction, since $f(\phi(\x_p^j))<0.$

\section{Comparison of Validation Accuracies}
\begin{figure}
\centering
 \includegraphics[width=0.8\linewidth]{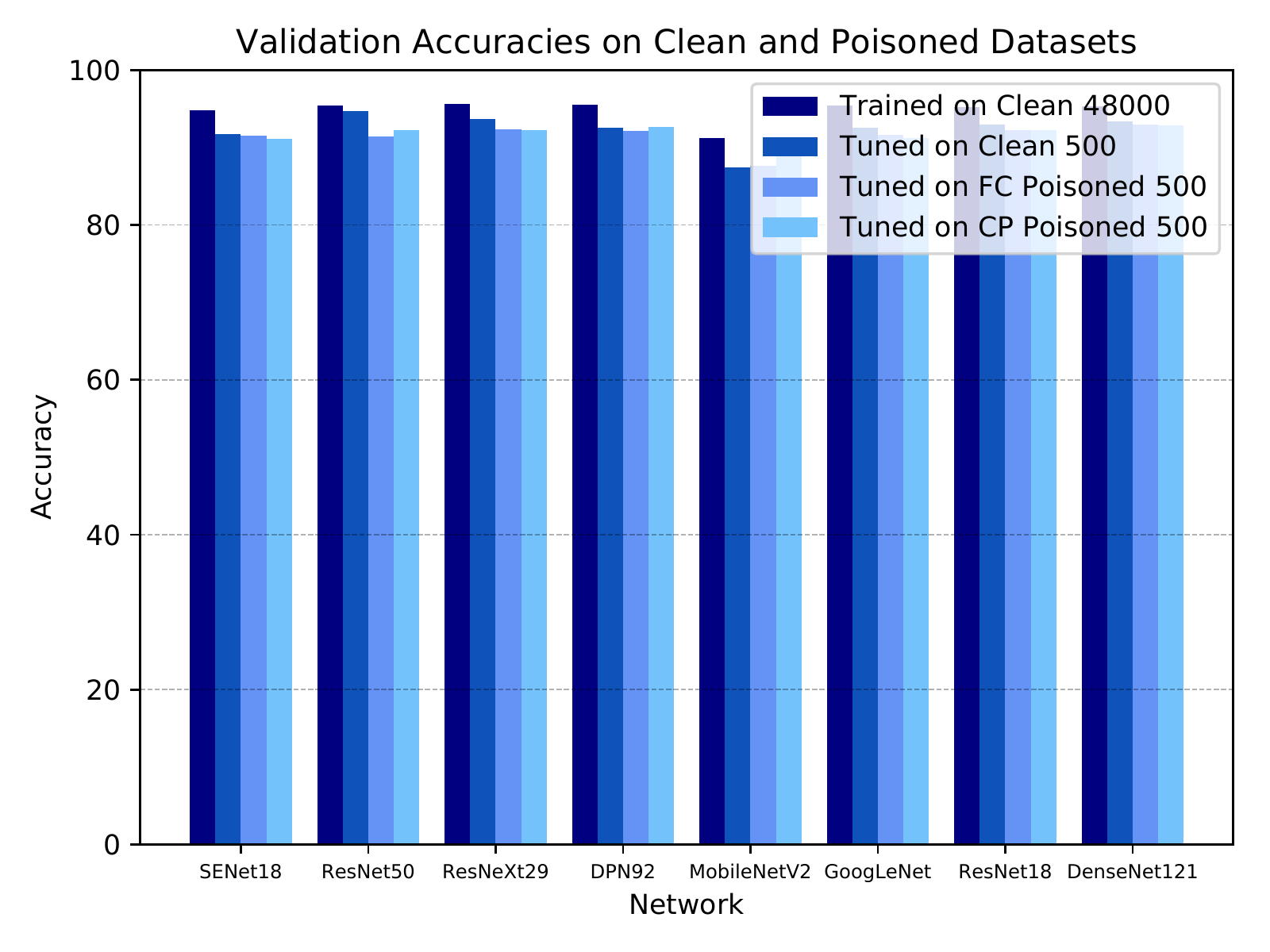}
 \caption{Accuracies on the whole CIFAR10 test for models trained or fine-tuned on different datasets. The fine-tuned models are initialized with the network trained on the first 4800 images of each class. There is little accuracy drop after fine-tuned on the poisoned datasets, compared with fine-tuning on the clean 500-image set directly. }
 \label{fig:compare_acc}
\end{figure}
To make data poisoning attacks undetectable, in addition to making the perturbations to nonobvious, the accuracy of the model fine-tuned on the poisoned dataset shall not drop too significantly, compared with fine-tuning on the same (except for the poisons) clean dataset. 
Figure~\ref{fig:compare_acc} shows that the drop in accuracy is indeed not obvious.

\section{Details of the qualitative example}
\begin{figure}
  \centering
  \includegraphics[width=.98\textwidth]{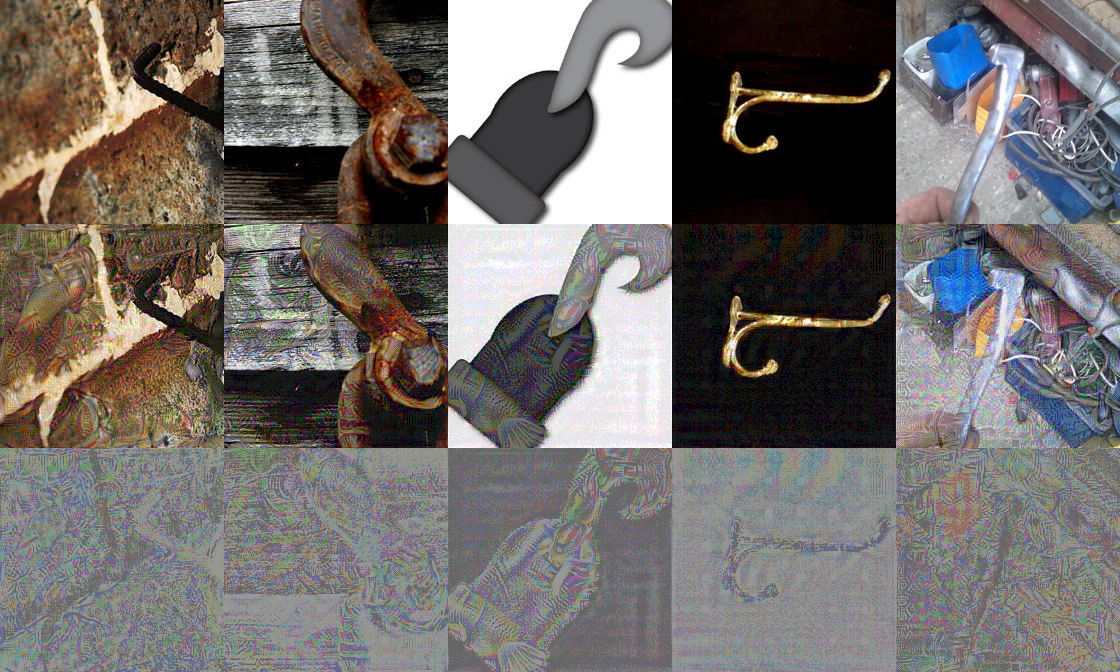}
  \includegraphics[width=.98\textwidth]{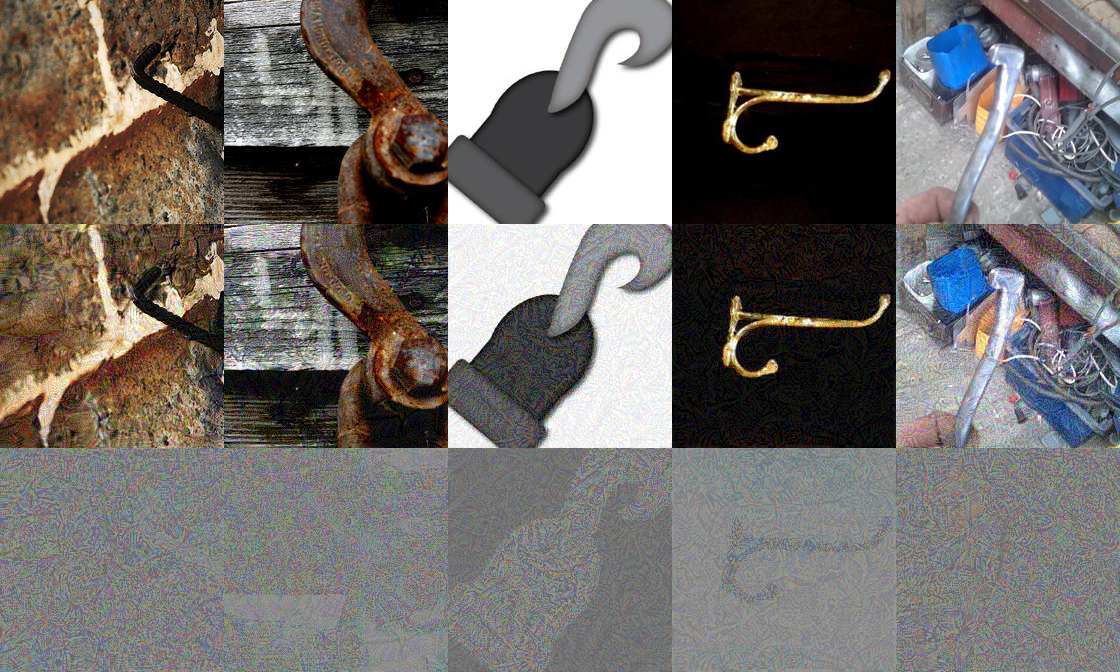}
  \caption{All the 5 poison images.}
  \label{fig:fishes}
\end{figure}
Both the target \emph{fish} image and the five \emph{hook} images used for crafting poisons come from the WebVision~\citep{li2017webvision} dataset, which has the same taxonomy as the ImageNet dataset.
Figure~\ref{fig:fishes} gives all the five poison examples.

\clearpage
\bibliography{example_paper}
\bibliographystyle{icml2019}

\end{document}